\documentclass[preprints,article,accept,moreauthors,pdftex]{Definitions/mdpi} 
\usepackage{nameref}
\usepackage{caption}
\usepackage[labelformat=simple]{subcaption}

\DeclareCaptionLabelFormat{subcaptionlabel}{\normalfont(\textbf{#2}\normalfont)}
\usepackage{epstopdf}
\firstpage{1} 
\pubvolume{22}
\issuenum{16}
\articlenumber{6088}
\pubyear{2022}
\copyrightyear{2022}
\externaleditor{}
\datereceived{ } 
\dateaccepted{} 
\datepublished{} 
\hreflink{} 

\Title{Blind-Spot Collision Detection System for Commercial Vehicles Using Multi Deep CNN Architecture}

\TitleCitation{Blind-Spot Collision Detection System for Commercial Vehicles Using Multi Deep CNN Architecture}

\Author{{Muhammad Muzammel} 
 $^{1,2}$, Mohd Zuki Yusoff $^{1,}$*, Mohamad Naufal Mohamad Saad $^{1}$, Faryal Sheikh $^{3}$ and Muhammad Ahsan Awais $^{1}$}

\AuthorNames{Muhammad Muzammel, Mohd Zuki Yusoff, Mohamad Naufal Mohamad Saad, Faryal Sheikh and Muhammad Ahsan Awais}

\AuthorCitation{Muzammel, M.; Yusoff, M.Z.;  Saad,  M.N.M.; Sheikh, F.; Awais, M.A.}

\address{%
$^{1}$ \quad Centre for Intelligent Signal \& Imaging Research (CISIR), Electrical and Electronic Engineering Department, Universiti Teknologi PETRONAS, {Seri Iskandar 32610}
, {Malaysia}\\ 
$^{2}$ \quad Laboratoire Images, Signaux et Systèmes Intelligents (LISSI), Université Paris-Est Créteil {(UPEC),} 
\mbox{94400 Vitry-sur-Seine, France}\\
$^{3}$ \quad Department of Management Sciences, {COMSATS University Islamabad, Lahore Campus},  
\mbox{Lahore {54000}, 
{Pakistan}}}

\corres{Correspondence: mzuki\_yusoff@utp.edu.my}

\abstract{Buses and heavy vehicles have more blind spots compared to cars and other road vehicles due to their large sizes. Therefore, accidents caused by these heavy vehicles are more fatal and result in severe injuries to other road users. These possible blind-spot collisions can be identified early using vision-based object detection approaches. Yet, the existing state-of-the-art vision-based object detection models rely heavily on a single feature descriptor for making decisions. In this research, the design of two convolutional neural networks (CNNs) based on high-level feature descriptors and their integration with faster R-CNN is proposed to detect blind-spot collisions for heavy vehicles. Moreover, a fusion approach is proposed to integrate two pre-trained networks (i.e., Resnet 50 and Resnet 101) for extracting high level features for blind-spot vehicle detection. The fusion of features significantly improves the performance of faster R-CNN and outperformed the existing state-of-the-art methods. Both approaches are validated on a self-recorded blind-spot vehicle detection dataset for buses and an online LISA dataset for vehicle detection. For both proposed approaches, a false detection rate (FDR) of 3.05\% and 3.49\% are obtained for the self recorded dataset, making these approaches suitable for real time applications.}


\begin{document}


\section{Introduction}
Although bus accidents are very rare around the globe, there are still approximately 60,000 buses are involved in traffic accidents in the United States every year. These accidents lead to 14,000~non-fatal injuries and 300 fatal injuries \cite{ref1}. 
Similarly, every year in Europe approximately 20,000 buses are involved in accidents that cause approximately 30,000 (fatal and non-fatal) injuries \cite{ref2}. These accidents mostly occurred due to thrill-seeking driving, speeding, fatigue, stress, and aggressive driver behaviors \cite{ref42,ref43}. Accidents involving buses and other road users, such as pedestrians, bicyclists, motorcyclists, or car drivers and passengers, usually cause more severe injuries to these road users \cite{ref3,ref4,ref5,ref6}.


{The collision detection systems of cars mostly focus on front and rear end collision scenarios \cite{ref13,ref14,ref15,ref16}. In addition, different drowsiness detection techniques have been proposed to detect car drivers’ sleep deprivation and prevent possible collisions \cite{abraham2018enhancing,shameen2018electroencephalography}. At the same time, buses operate in a complicated environment where a significant number of unintended {obstacles} such as pulling out from bus stops, {passengers} unloading, pedestrians crossing in front of {buses}, and bus stop structures, etc. \cite{mcneil2002performance,pecheux2016test,wei2014task}, are present. Additionally, buses have higher chances of side collisions due to constrained spaces and maneuverability~\cite{mcneil2002performance}. Especially at turns, researchers found that the task demand on bus drivers is very high~\mbox{\cite{pecheux2016test,wei2014task}.} }

{Further, heavy vehicles and buses, which have more blind spots compared to cars and other road users in these environments, are at higher {risks} of collisions \cite{ref7, ref8, ref40}. Improvements to heavy vehicle and bus safety have been initiated by many countries through the installation of additional mirrors. Yet, there are still some blind-spot areas where drivers cannot see other road users \cite{ref9, ref10}. In addition, buses may have many passengers on board. A significant number of on-board passenger incidents have been reported due to sudden braking or stopping \cite{zhang2000develop}. These challenges may entail different requirements for collision {detections} for public/transit buses than for cars. A blind-spot collision detection system can be designed for buses to predict impending collisions in their proximity and to reduce operational interruptions. It could provide adequate time for the driver to smoothly push the brake or take any other precautionary measures to avoid such imminent collision threats as well as avoid {injuries and trauma inside the bus.}}

\textls[+15]{Over the past few years, many types of collision detection techniques have been proposed~\cite{ref11,ref12,ref13,ref14,ref41}.} Among these, vision-based collision detection techniques provide reliable detection of vehicles across a large area \cite{ref13,ref14,ref41}. This is due to cameras that provide a wide field of view.
Several vision-based blind-spot collision detection techniques for cars and other vehicles have been proposed \cite{ref13,ref14,ref15,ref16,ref17,ref18,ref19}. In vision-based techniques, the position of the camera plays a significant role. Depending on the position of the installed camera, vision-based blind-spot collision detection systems are categorized as rear camera based~\cite{ref15,ref20} or side camera-based systems \cite{ref17,ref18,ref19,ref41}. Rear camera-based vision systems detect vehicles by acquiring the images from a rear fish-eye camera. The major drawback of using a rear fish-eye camera is that the captured vehicle suffers from severe radial distortions, leading to huge differences in appearance for different positions \cite{ref15}.

In contrast, side camera-based vision systems have the camera installed directly at the bottom or next to the side mirrors that directly face the blind spot and detect the approaching vehicles. In these systems, the vehicle appearance drastically changes with its position; yet, it has the advantage of high resolution images for vehicle detection \cite{ref15}.

In vision-based blind-spot vehicle techniques, deep convolutional neural network (CNN) models often achieve better performance \cite{ref14,ref16} compared to conventional machine learning models (based on the appearance, histogram of oriented gradients (HOG) features, etc.) \cite{ref15,ref17,ref18}. This is due to convolutional layers that can extract and learn more pure features from the raw RGB channels than traditional algorithms such as HOG. However, blind-spot vehicle detection is still challenging on account of the large variations in appearance and structure, especially ubiquitous occlusions that further increase the intra-class~variations.

Recently, deep learning techniques prove to be a game changer in object detection. Many deep learning models have been proposed to detect different types and sizes of objects in images \cite{ref29,ref32,ref33}. Among these models, two-stage object detectors show better accuracy compared to one-stage object detectors \cite{lin2017focal, huang2017speed,du2020overview}. Therefore, two-stage object detectors, such as faster R-CNN \cite{ref29}, seem to be more suitable for blind-spot vehicle detection. In faster R-CNN, a self-designed CNN or a pre-trained network (such as VGG16, ResNet50, and ResNet-101, etc.) is used to extract a feature map \cite{theckedath2020detecting,he2016deep}. These networks are trained on a large dataset and are proven to be better in performance compared to simple convolutional neural networks (CNNs). In medical applications, it has been reported that multi-CNNs performed much better in residual feature extraction and classification compared to single CNNs \cite{yang2017multimodal, MUZAMMEL2021106433,mendels2017hybrid}.        

In this paper, we propose a novel blind-spot vehicle detection technique for commercial vehicles based on multi convolutional neural networks (CNNs) and faster R-CNN. Two different convolutional neural network-based approaches/models with faster R-CNN as an object detector are proposed for blind-spot vehicle detection. In the first approach/model, two self designed CNNs networks are used to extract the features, and their outputs are concatenated and fed to another self designed CNN. Next, faster R-CNN uses these high-level features for vehicle detection. In the second approach/model, two ResNet CNN networks (ResNet-50 and ResNet-101) are concatenated with the self-designed CNN to extract features. Finally, these extracted features are fed to the faster R-CNN for blind-spot vehicle detection. The scientific contributions of this research are as follows:
\begin{enumerate}
\item	Design of two high-level CNN based feature descriptors for blind-spot vehicle detection for heavy vehicles;
\item	Design of fusion technique for different high level feature descriptors and its integration with the faster R-CNN. In addition, performance comparison with existing state-of-the-art approaches;
\item   Introduction of fusion technique for pre-trained high-level feature descriptors for object detection application.
\end{enumerate}

\section{Related Work}

The recent deep convolutional neural network (CNN) based algorithms depict extraordinary performance in various vision tasks \cite{ref21,ref22,ref23,ref24}. Convolutional neural networks extract features from the raw images through a large amount of training with high flexibility and generalization capabilities. The first CNN based object detection and classification system was presented in 2013 \cite{ref25,ref26}. Up to now, many deep learning-based object detection and classification models have been proposed, including region based convolutional neural network (R-CNN) \cite{ref27}, fast R-CNN~\cite{ref28}, faster R-CNN \cite{ref29}, single shot multibox detector (SSD) \cite{ref30}, R-FCN \cite{ref31}, you only look once (YOLO) \cite{ref32}, and YOLOv2 \cite{ref33}.

R-CNN models achieve promising detection performance and are a commonly employed paradigm for object detection \cite{ref27}. They have essential steps, such as object regional proposal generation with selective search (SS), CNN feature extraction, selected objects classification, and regression based on the obtained CNN features. However, there are large time and computation costs to train the network due to repeated extraction of CNN features for thousands of object proposals \cite{ref34}.

In fast R-CNN \cite{ref28}, the feature extraction process is accelerated by sharing the forward pass computation. Due to the regional proposal generation by selective search (SS), it still appears to be slow and requires significant computational capacity to train it. In faster R-CNN \cite{ref29}, “regional proposal generation using SS” was replaced by the "proposal generation using CNN”. This increases the computational capacity of the network and makes it efficient and quick compared to the R-CNN and fast R-CNN.

YOLO \cite{ref32} frame object detection is a regression problem to separate bounding boxes and associated class probabilities. In YOLO, a single CNN predicts the bounding boxes and class probabilities for these boxes. It utilizes a custom network based on the GoogLeNet architecture. An improved model called YOLOv2 \cite{ref33} achieves comparable results on standard tasks. YOLOv2 employs a new model called Darknet-19, which has 19~convolutional layers and 5 max-pooling layers. This new model only takes 5.58 s to compute results. However, the YOLOv2 network still lacks some important elements, it has no residual blocks, no skip connections, and no up-sampling, etc. 

The YOLOv3 network is the advanced version of YOLOv2 and incorporates all of these important elements. YOLOv3 is a 53 layer network trained on Imagenet. For object detection, YOLOv3 has 53 more layers stacked onto it and gives us a 106 layer fully convolutional underlying architecture \cite{ref44}. Recently, two new versions of YOLO were introduced, named YOLOv4 and YOLOv5, respectively \cite{yolov4,yolov5}. Other than YOLO, there are also other one-stage object detectors, such as SSD \cite{ref30} and RetinaNet \cite{lin2017focal}. 

Recent studies show that two-stage object detectors obtained better accuracy compared to one-stage object detectors \cite{lin2017focal, huang2017speed}, thus, making faster R-CNN a suitable candidate for blind-spot vehicle detection. However, in these object detectors, the whole system accuracy is profoundly dependent on the feature set obtained from the neural networks. In recent object detectors, it has also been proposed to collect features from different stages of the neural network to improve the system performance \cite{lin2017feature,tan2020efficientdet}. In medical applications, it has been demonstrated that the usage of multiple feature extractors can significantly improve system accuracy \cite{yang2017multimodal, MUZAMMEL2021106433,mendels2017hybrid}. 

Thus, to increase system accuracy, in this research multiple CNN networks based blind-spot vehicle detection approaches are proposed. Along with the fusion of a self-designed convolutional neural network, system performance is also investigated using a fusion approach for pre-trained convolutional neural networks.

\section{Proposed Methodology}
\begin{figure}[H]
\includegraphics[width=13cm]{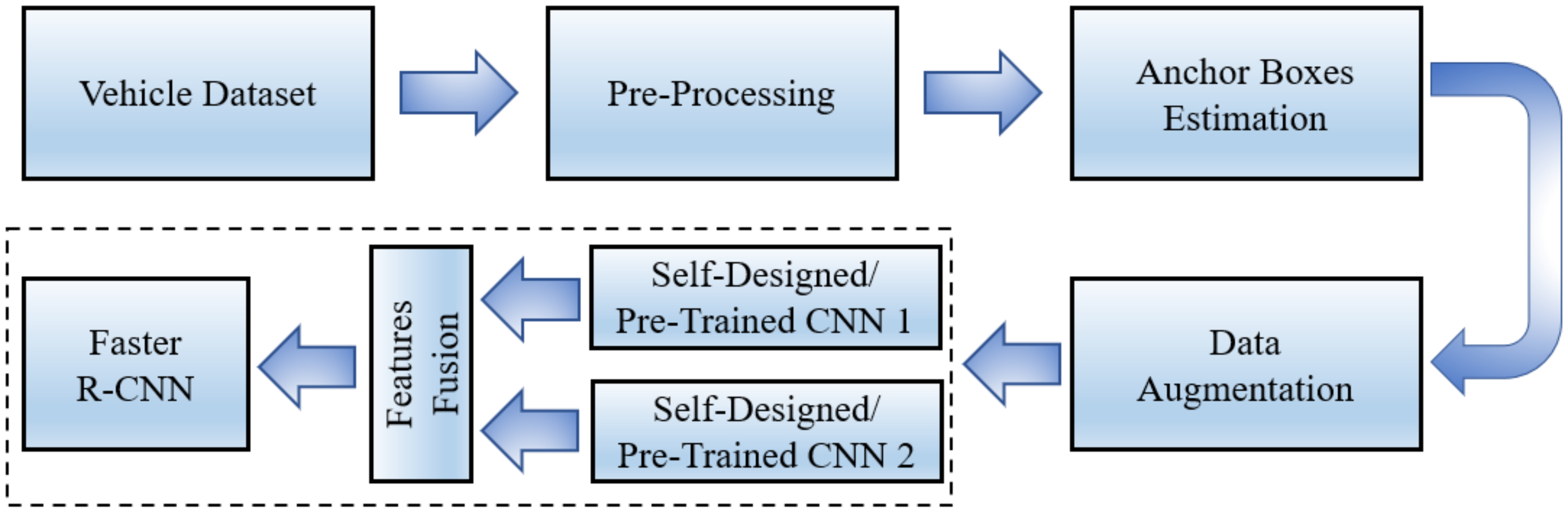}
\caption{{Steps of the proposed approaches to detect blind-spot vehicles using faster R-CNN object~detection.} \label{method1}}
\end{figure}

The proposed methodology comprises several steps, including pre-processing of datasets, anchor boxes estimation, data augmentation, and multi CNN network design, as shown in Figure \ref{method1}.

\subsection{Pre-Processing}
For the self-recorded dataset, image labels were created using MATLAB 2019a “Ground Truth Labeller App”, whereas for the online dataset, ground truths were provided with the image set. Next, images were resized to {224 $\times$ 224 $\times$ 3} 
to enhance the computation performance of the proposed deep neural networks.

\subsection{Anchor Boxes Estimation}
\begin{figure}[H]
\includegraphics[width=11cm]{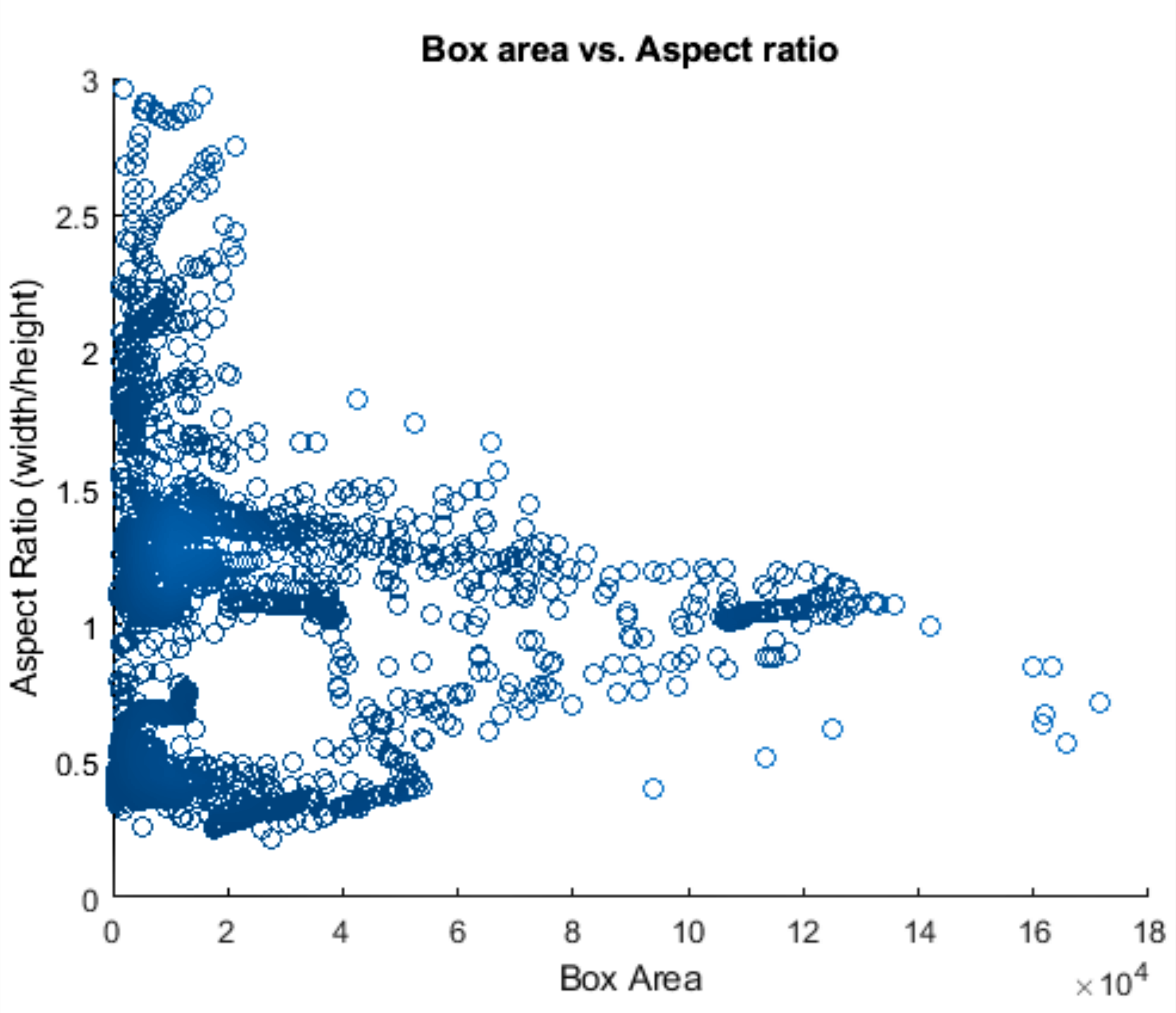}
\caption{{Anchor} 
 boxes plot to identify sizes and shapes of different vehicles for faster R-CNN object detection. {Each blue circle indicates the label box area versus the label box aspect ratio.}} 
 \label{anchorbox}
\end{figure}

Anchor boxes are important parameters of deep learning object recognition. The shape, scale, and the number of anchor boxes impact the efficiency and accuracy of the object detector. Figure \ref{anchorbox} indicates the plot of aspect ratio and box area of the self-recorded dataset.

\vspace{-6pt}

The anchor boxes plot reveals that many vehicles have a similar size and shape. However, vehicle shapes are still spread out, indicating the difficulty of choosing anchor boxes manually. Therefore, a clustering algorithm presented in \cite{ref33} was used to estimate anchor boxes. It groups similar boxes together using a meaningful metric.



\subsection{Data Augmentation}
In this work, data augmentation is performed to minimize the over-fitting problem and to improve the proposed network's robustness against noise. Random brightness augmentation technique is considered to perturb the images. The brightness of the images is augmented by randomly darkening and brightening the images. The darkening and brightening values randomly range from [0.5, 1.0] and [1.0, 1.5], respectively.

\subsection{Proposed CNNs and Their Integration with Faster R-CNN}
\label{NA}

\begin{figure}[H]
\includegraphics[width=13cm]{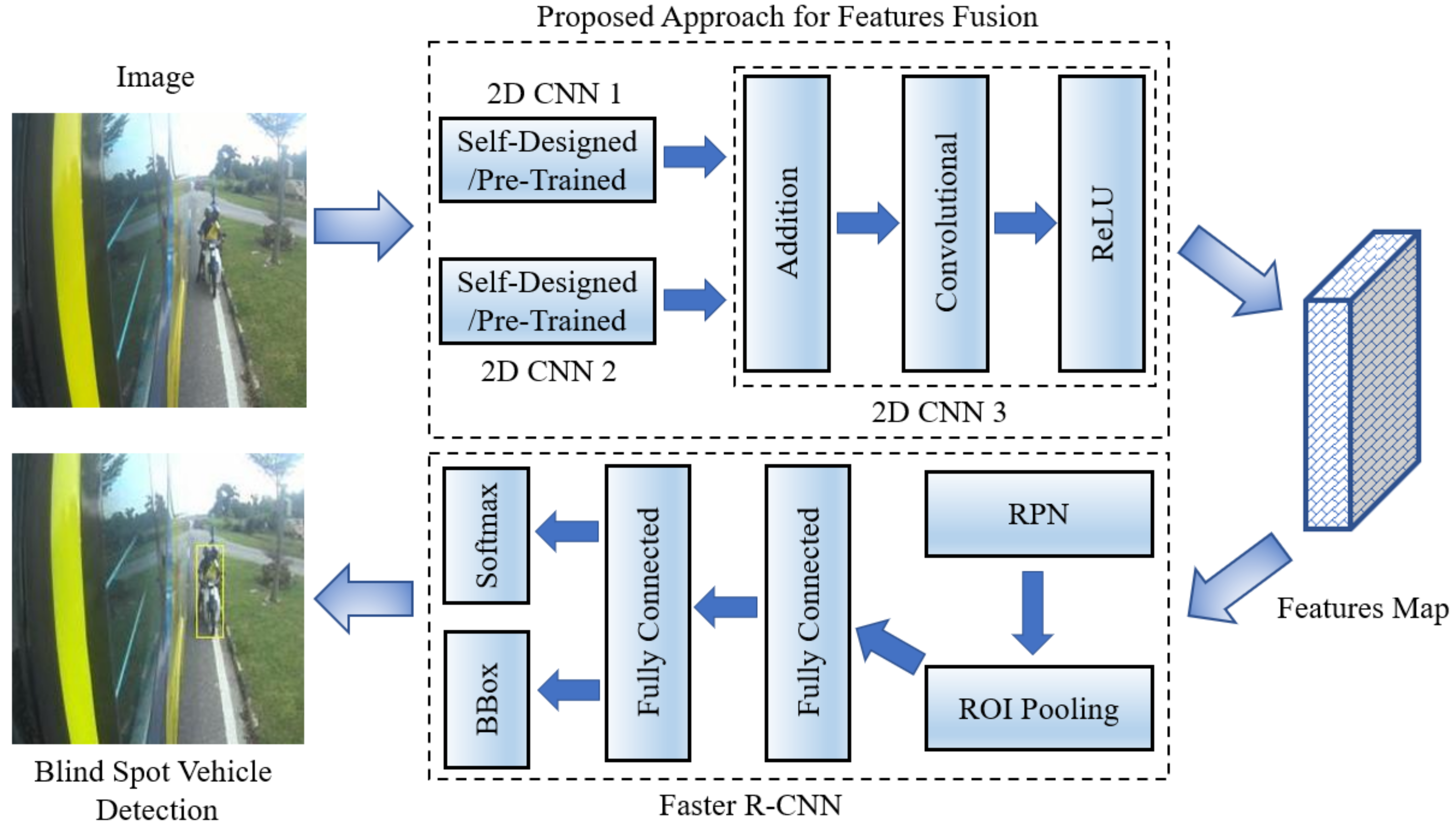}
\caption{\textls[-15]{{Layer wise integration of proposed models with faster R-CNN for blind-spot \mbox{vehicle~detection}}.} \label{Myapproach1}}
\end{figure}

Initially, the same images are fed to two different deep learning networks to extract high-level features. Subsequently, these high-level features are fed to another CNN architecture to combine and smooth these features. Finally, faster R-CNN based object detection is performed to detect impending collisions. The layer wise connection of deep learning architectures and their integration with faster R-CNN are shown in Figure \ref{Myapproach1}.

\subsubsection{Proposed High Level Feature Descriptors Architecture} 
\begin{figure}[H]
\includegraphics[width=10cm]{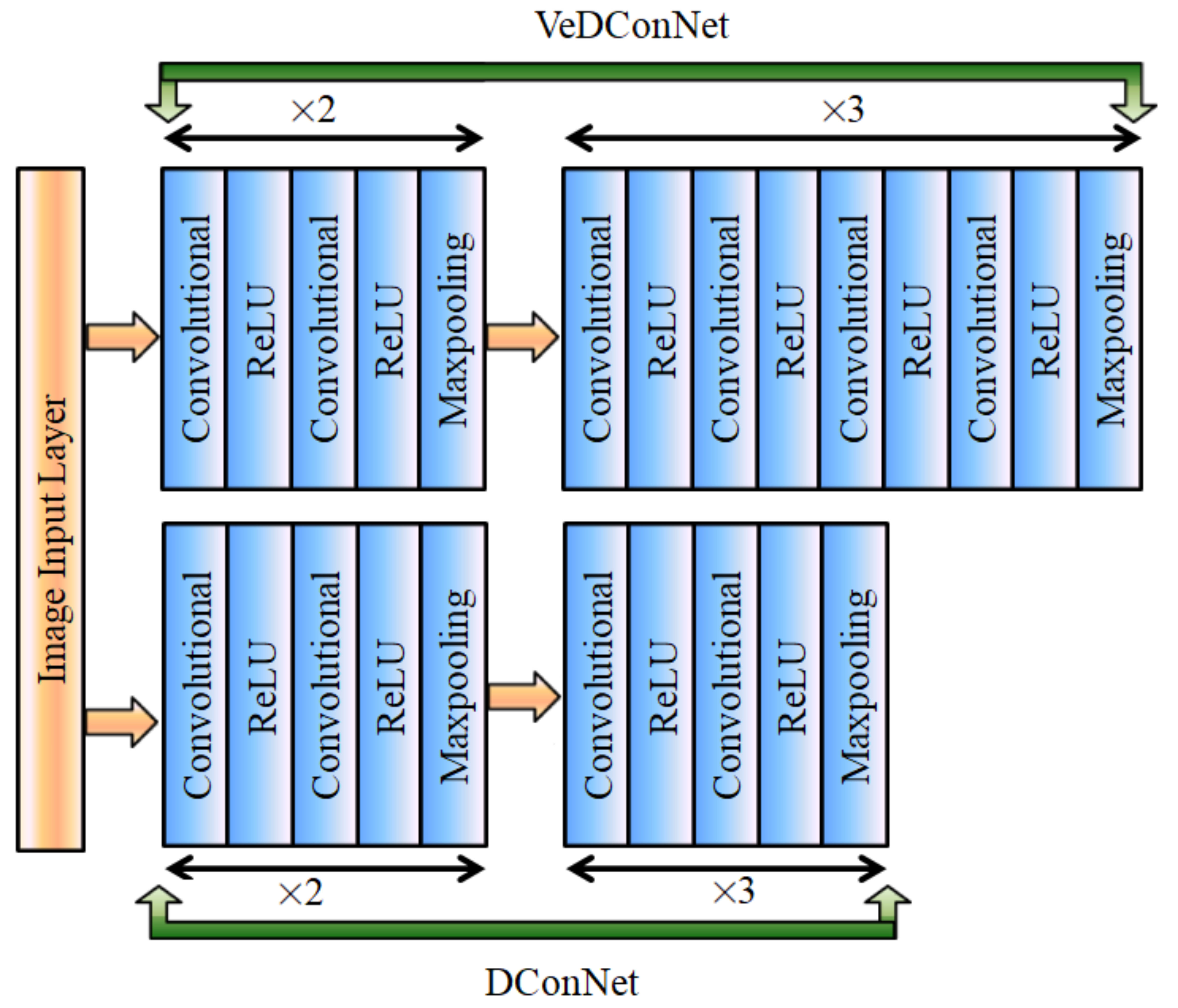}
\caption{Proposed 2D CNN architectures to extract deep features for blind-spot vehicle {detection.} 
 \label{Myapproach2}}
\end{figure}

Two different approaches are used to extract deep features: (1) self-designed convolutional neural networks and (2) pre-trained convolutional networks, as shown in Figure \ref{Myapproach1}. Additional details of these feature descriptors are given below.    
\paragraph{Self-Designed High-Level Feature Descriptors}
\label{NA:Exp1}
In first approach, multiple self-designed convolutional neural networks are connected with the faster R-CNN network. The layer wise connection of two self-designed CNN networks (named DConNet and VeDConNet) is shown in Figure \ref{Myapproach2}. Initially, DConNet and VeDConNet are used to extract deep features, and their output is provided to the third 2D CNN architecture for the purpose of features addition and smoothness.

Both DConNet and VeDConNet architectures consist of five convolutional blocks. In DConNet, all five blocks are composed of a two 2D convolutional and ReLU layers. In addition, at the end of each block there is a max-pooling layer. In VeDConNet, the initial two blocks are similar to DConNet as they consist of two 2D convolutional layers, each followed by a ReLU activation function, where a max-pooling layer is also available after the second ReLU activation function. The other three blocks of VeDConNet comprise four convolutional layers, each followed by the ReLU layer and maxpooling layer after the fourth ReLU activation~function.

\begin{figure}[H]
\includegraphics[width=9cm]{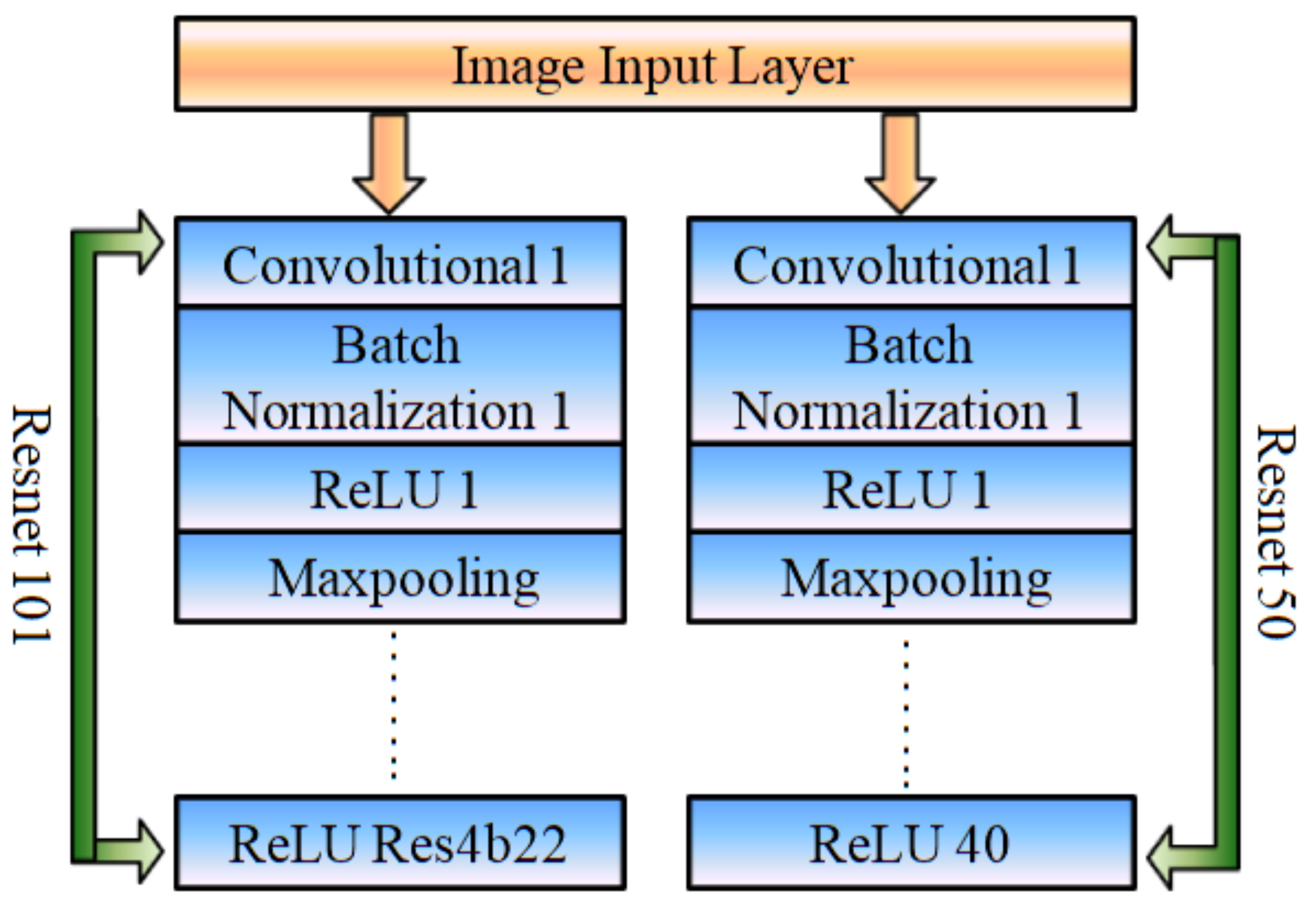}
\caption{Pre-trained Resnet 50 and Resnet 101 networks for extracting deep features. \label{Myapproach3}}
\end{figure}

\paragraph{Pre-Trained Feature Descriptors}
In the second approach, two pre-trained convolutional networks (i.e., Resnet 101 and Resnet 50) are linked with the third CNN architecture, which is further connected with the faster R-CNN network for the purpose of vehicle detection. The features obtained from ReLU Res4b22 and ReLU 40 layers of ResNet 101 and ResNet 50, respectively, as given in Figure \ref{Myapproach3}. 

\subsubsection{Features Addition and Smoothness}
The high level features obtained from the two self-designed/pre-trained CNN architectures are added together through the addition layer, as shown in Figure \ref{Myapproach1}. Let $F_{1}(x)$ and $F_{2}(x)$ be the output of the first and second deep neural networks, then their addition $H(x)$ is given as:    

\begin{linenomath}
\begin{equation}
H(x) = F_{1}(x) + F_{2}(x)
\end{equation}
\end{linenomath}

The addition layer is followed by the convolutional layer and ReLU activation function for the features smoothness. 

\subsubsection{Integration with Faster R-CNN}
As shown in Figure \ref{Myapproach1}, faster R-CNN takes high level features from the ReLU layer to perform the blind-spot vehicle detection. The obtained features map is fed to region proposal network (RPN) and ROI pooling layer of the faster R-CNN.  
The loss function of faster R-CNN can be divided into two parts: R-CNN loss \cite{ref28} and RPN loss \cite{ref29}, which is shown in the equations below:

\begin{linenomath}
\begin{equation}
L(p,u,t^u,v) = L_{cls}(p,u) + \lambda{'} L_{reg}(t^u,v)
\end{equation}
\end{linenomath}

\begin{linenomath}
\begin{equation}
L(\{p_i\},\{t_i\}) = \frac{1}{L_{cls}}\sum_{i}L_{cls}(p_i,p_i^*) + \lambda\frac{1}{L_{reg}}\sum_{i}p_i^*L_{reg}(t_i,t_i^*)
\end{equation}
\end{linenomath}

 The detailed description of the faster R-CNN architecture and the above equations is given in references \cite{ref28,ref29,ref39}.


\section{Results and Discussion}
In this section, the vehicle detection using the proposed deep learning models is discussed in detail. We compared the performance of both approaches with each other and with the state-of-the-art benchmark approaches. This section also includes the dataset description along with the details of the proposed network implementation.

\subsection{Dataset}
\begin{figure}[H]
\includegraphics[width=12cm]{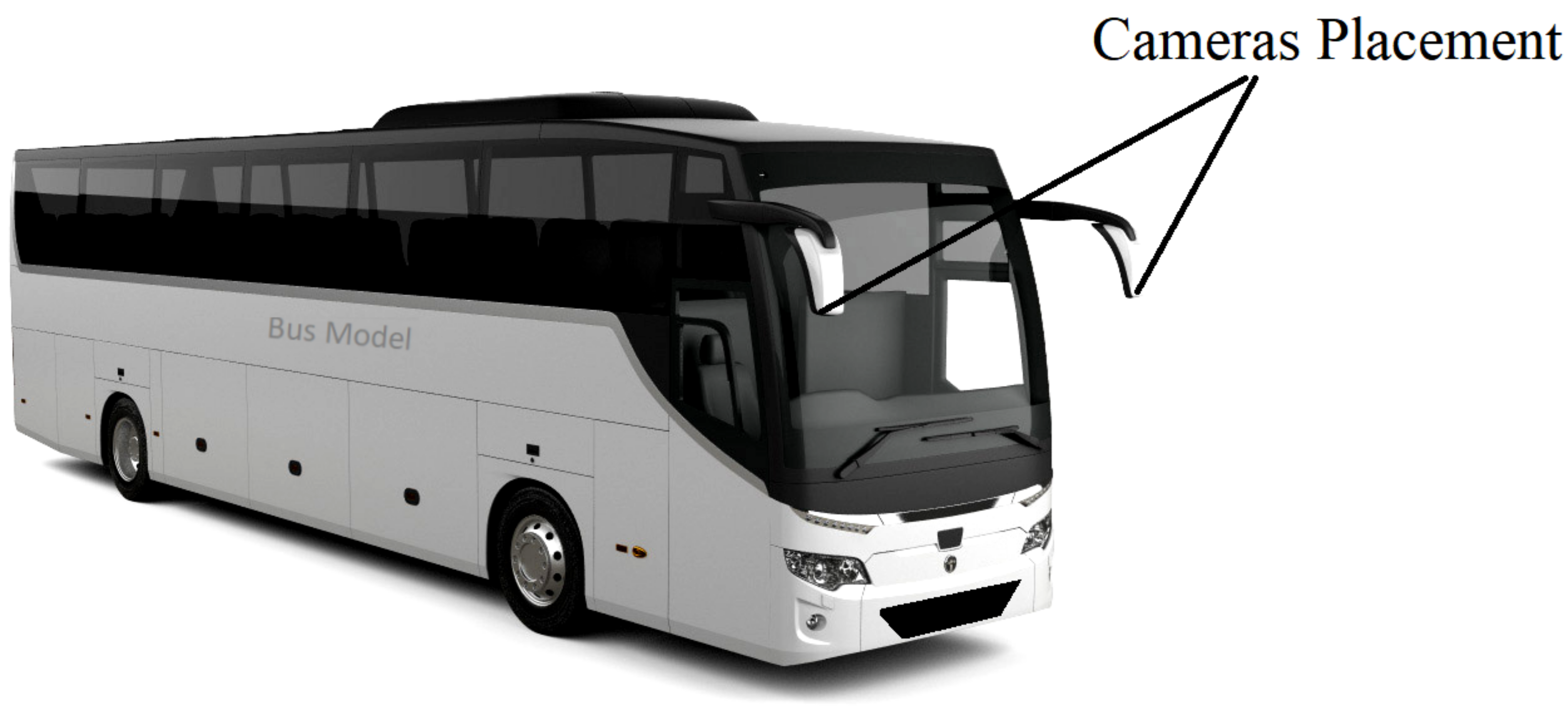}
\caption{Cameras mounted on the bus mirrors to detect the presence of vehicles in blind spots. \label{bus}}
\end{figure}

A blind-spot collision dataset was recorded by attaching cameras to the side mirrors of a bus. The placement of cameras is shown in Figure \ref{bus}.

The dataset was recorded in Ipoh, Seri Iskandar and along Ipoh-Lumut highway in Perak, Malaysia. Ipoh is a city in northwestern Malaysia, whereas Seri Iskandar is located about 40 km southwest of Ipoh. Universiti Teknologi PETRONAS  is also located in the new township of Seri Iskandar. Data were recorded in multiple round trips from Seri Iskandar to Ipoh for different lighting conditions. In addition, data were recorded in the cities of Ipoh and Seri Iskandar for dense traffic scenarios. Moreover, Malaysia has a tropical climate and the rainfall remains high year-round, thus allowing us to easily record data in different weather conditions. Finally, a set of 3000 images from the self-recorded dataset was selected in which vehicles appeared in blind-spot areas. 

To the best of our knowledge, there is no publicly available online dataset for heavy vehicles. Therefore, a publicly available online dataset named “Laboratory for Intelligent and Safe  Automobiles (LISA)” \cite{ref35} for cars was used to validate the proposed method. In the LISA dataset, the camera was installed at the front of the car. The detailed description of both datasets is given in Table \ref{tabdata}. Both datasets are divided randomly into 80\% for training and 20\% for testing.

\begin{table}[H] 
\caption{Utilized dataset characteristics to validate the proposed deep CNN based approaches for blind-spot collision detection.\label{tabdata}}
\setlength{\cellWidtha}{\textwidth/4-2\tabcolsep-0in}
\setlength{\cellWidthb}{\textwidth/4-2\tabcolsep--0.4in}
\setlength{\cellWidthc}{\textwidth/4-2\tabcolsep-0in}
\setlength{\cellWidthd}{\textwidth/4-2\tabcolsep-0.4in}
\scalebox{1}[1]{\begin{tabularx}{\textwidth}{>{\centering\arraybackslash}m{\cellWidtha}>{\centering\arraybackslash}m{\cellWidthb}>{\centering\arraybackslash}m{\cellWidthc}>{\centering\arraybackslash}m{\cellWidthd}}
\toprule
\textbf{Dataset}	& \textbf{Data Description}	& \textbf{Source of Recording} & \textbf{Total Images}\\
\midrule
Self-Recorded Dataset for Blind Spot Collision Detection & Different road scenarios with multiple vehicles and various traffic and lighting conditions.	& Bus   & 3000 \\
LISA-Dense \cite{ref35} & Multiple vehicles, dense traffic, daytime, highway.	& Car & 1600\\
LISA-Sunny \cite{ref35} & Multiple vehicles, medium traffic, daytime, highway.	& Car & 300\\
LISA-Urban \cite{ref35} &Single vehicle, urban scenario, cloudy morning.	& Car & 300\\
\midrule
Total  & &  & 5200\\
\bottomrule
\end{tabularx}}
\end{table}


\subsection{Network Implementation Details}
\textls[-5]{The proposed work was implemented on the Intel® Xeon(R) E-2124G CPU @ 3.40~GHz (installed memory 32~GB), with a NVIDIA Corporation GP104GL [Quadro P4000] graphics card. MATLAB 2019a was used as platform to investigate the proposed methodology. }

In the first approach, both CNN based feature extraction architectures (i.e., DConNet and VeDConNet) have five blocks with N number of convolutional filters for each block. Therefore, the number of convolutional filters of the five blocks from the input to the output is equal to N = [64, 128, 256, 512, 512]. Moreover, after the addition layer, there was also convolution layer with a total of 512 filters. For all these convolutional layers, the filter size was 3 $\times$ 3, and ReLU was used as an activation function. At the same time, the stride and the pool size of the max-pooling layer was 2 $\times$ 2.

In the second approach, for Resnet 101 and Resnet 50, standard weights were used. Moreover, after the addition layer, there was a convolution layer with a total of 512 filters and ReLU as an activation function.  

In both approaches, we used an SGDM optimizer with a learning rate of 10$^{-3}$ and a momentum of 0.9. The batch size was set to 20 samples, and the verbose frequency was set to 20. Negative training samples are set equal to the samples that overlap with the ground truth boxes by 0 to 0.3. However, positive training samples are set equal to the samples that overlap with the ground truth boxes by 0.6 to 1.0.


\subsection{Evaluation Matrix}

The existing state-of-the-art approaches measure the performance in terms of true positive rate (TPR), false detection rate (FDR), and frame rate \cite{ref35,ref36,ref37,ref38}. Therefore, the same parameters are used to evaluate the performance of the proposed models. TPR (also known as sensitivity) is the ability to correctly detect blind-spot vehicles. FDR refers to the false blind-spot vehicle detection among the total detection incidents. Moreover, the frame rate is defined as the total number of frames processed in one second \cite{ref36}. If {TP}, {FN}, and {FP} 
 represent the true positive, false negative, and false positive, respectively, then the formulas for TPR and FDR are given as:
 
\begin{linenomath}
\begin{equation}
\text{TPR} (\%)=  {\frac{\text{TP}}{\text{TP + FN}}} \times 100
\end{equation}
\end{linenomath}

\begin{linenomath}
\begin{equation}
\text{FDR} (\%)=  {\frac{\text{FP}}{\text{TP + FP}}} \times 100
\end{equation}
\end{linenomath}


\subsection{Results Analysis}
The proposed approaches/models appeared to be successful in detecting the vehicles for both self-recorded and online datasets. A few of the images from blind-spot detection are shown in Figure \ref{moto}. 



\begin{figure}[H]
    \begin{subfigure}[t]{0.5\textwidth}
        \includegraphics[width=0.88\textwidth]{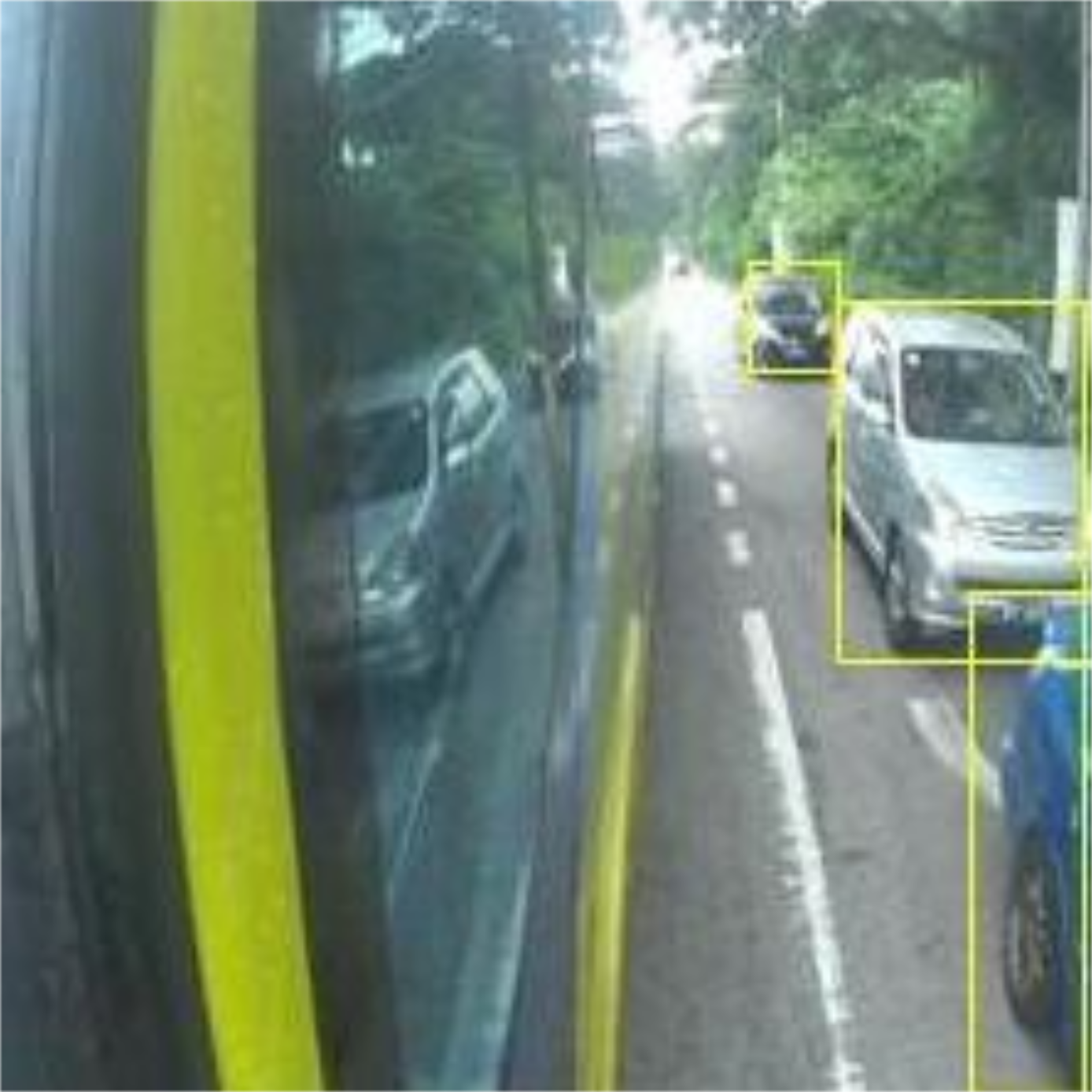}
        \caption{\centering} 
        \label{moto1}
    \end{subfigure}
\vspace{6pt}
    \begin{subfigure}[t]{0.5\textwidth}
        \includegraphics[width=0.88\textwidth]{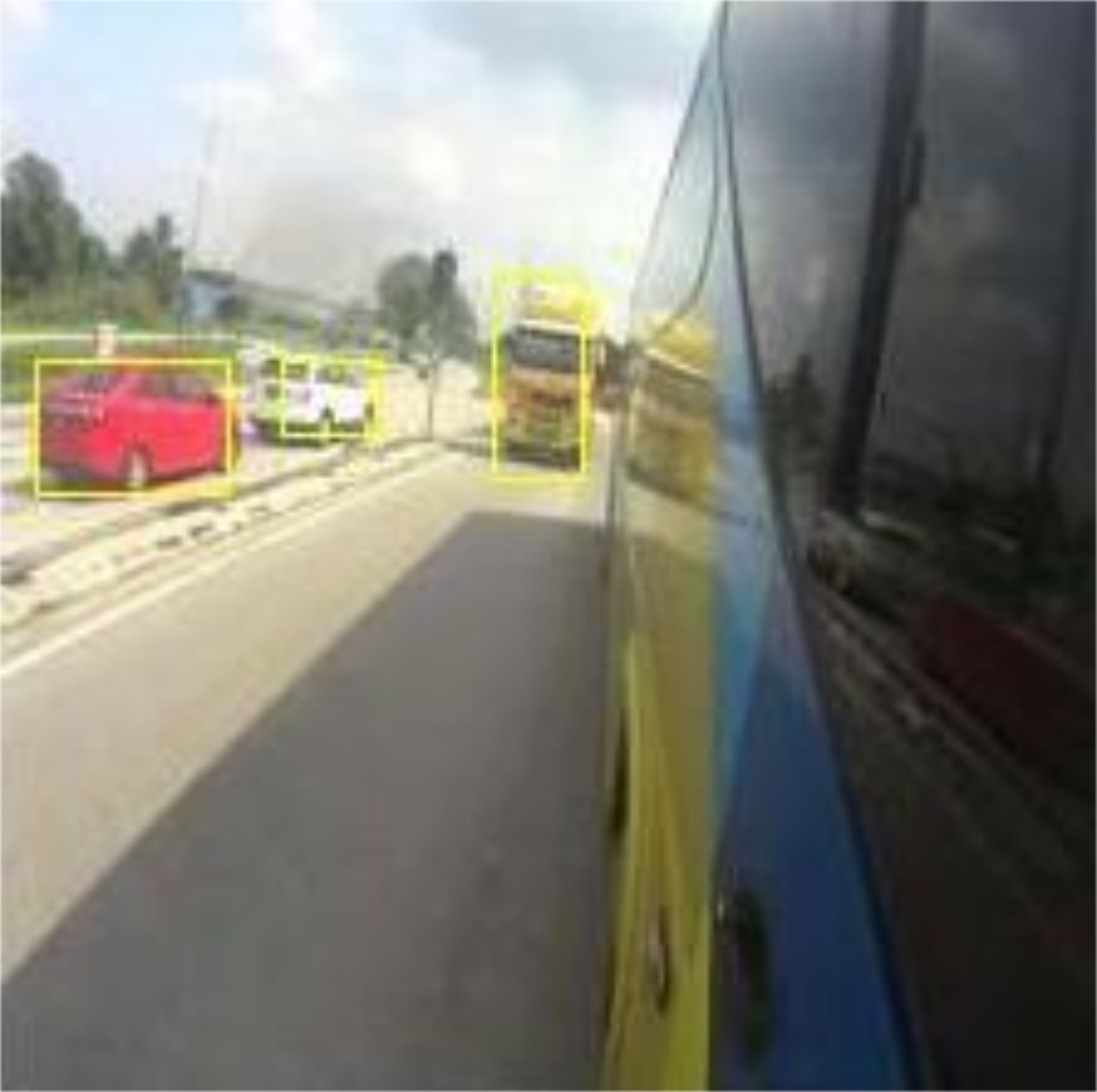}
        \caption{\centering} 
        \label{moto2}
    \end{subfigure}
\vspace{6pt}
    \begin{subfigure}[t]{0.5\textwidth}
        \includegraphics[width=0.88\textwidth]{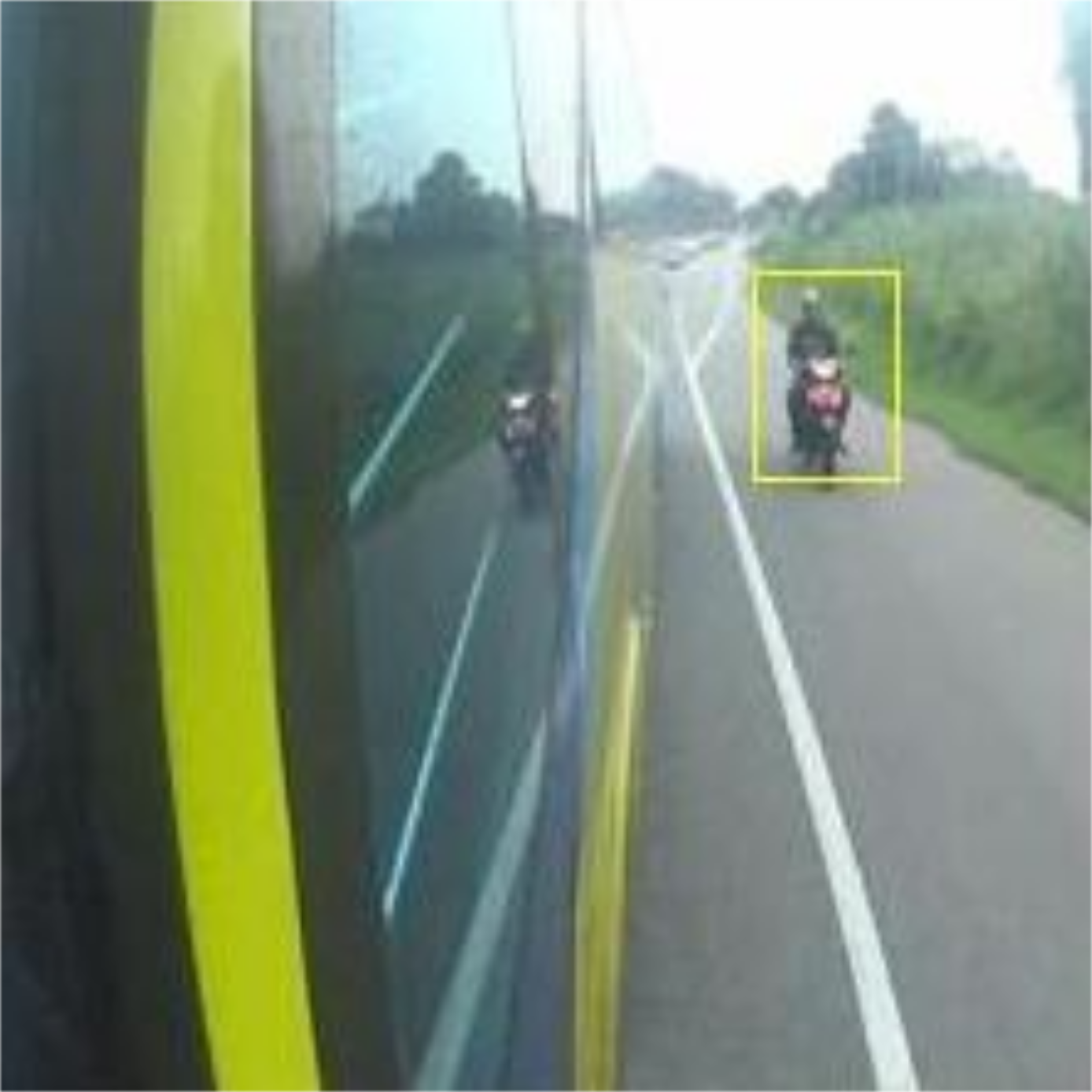}
        \caption{\centering} 
        \label{moto3}
    \end{subfigure}
\vspace{6pt}
    \begin{subfigure}[t]{0.5\textwidth}
        \includegraphics[width=0.88\textwidth]{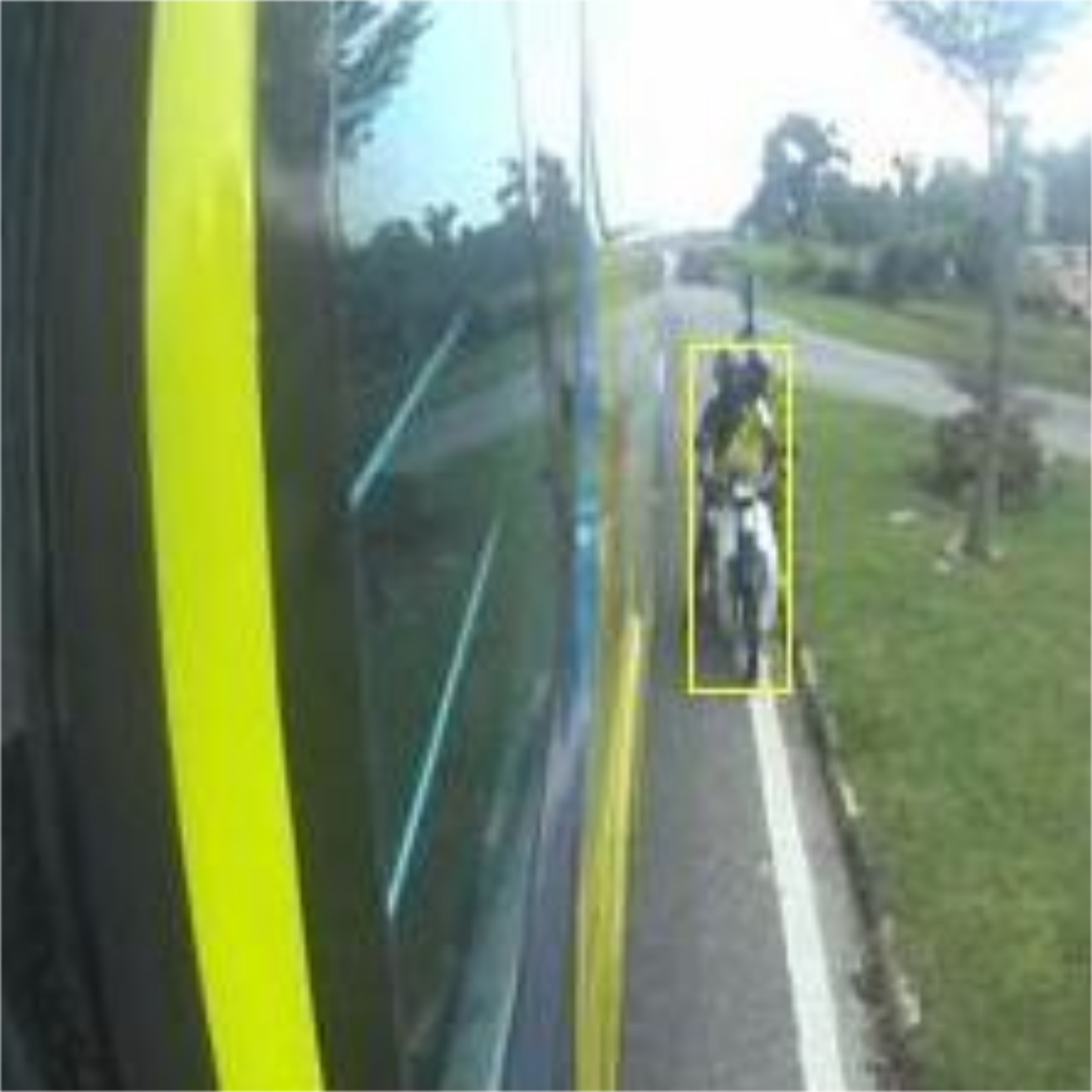}
        \caption{\centering} 
        \label{moto4}
    \end{subfigure}
\caption{Different types of vehicle detection from self-recorded dataset: (\textbf{a}) three different vehicles in a parallel lane with the bus; (\textbf{b}) one truck in a parallel lane and two cars in the opposite lane; (\textbf{c})~motorcycle at a certain distance; (\textbf{d}) motorcycle very close to the bus. \label{moto}}
\end{figure} 


Figure \ref{moto} shows that the proposed CNN based models were successfully able to detect different types of vehicles, including light and heavy vehicles and motor bikes, in different scenarios and lighting conditions. The proposed work was successful enough to recognize multiple vehicles simultaneously, as shown in Figure \ref{moto}a,b. These figures also show the presence of shadows along with the vehicles. It reveals the significance of the proposed vehicle detection algorithm, as it was capable of differentiating remarkably between real vehicles and their shadows; this leads to the notable reduction of possible false detection. 

Furthermore, it is shown in Figure \ref{moto}c,d that the proposed technique detects a motorcyclist approaching and driving very close to the bus. A small mistake by the bus driver in such scenarios could lead to a fatal accident. Therefore, the blind-spot collision detection systems are very important for heavy vehicles.

Similarly, vehicle detection from the online LISA dataset \cite{ref35} is shown in Figure~\ref{car}.
From Figure \ref{car}, our models were successfully able to detect all types of vehicles in different scenarios using the LISA dataset. Figure \ref{car}a,b show the detection of vehicles in dense scenarios. The proposed models were reliable enough to detect multiple vehicles simultaneously in dense scenarios, even in the presence of vehicle shadows on the road. \mbox{Figure \ref{car}c,d} exhibit the detection of vehicles on a highway, and Figure \ref{car}e,f convey the detection of vehicle in urban areas. In both figures, we can see the presence of lane markers on the road, which were successfully neglected by the proposed systems. Furthermore, \mbox{Figure \ref{car}f} shows a person crossing the road; this could lead to a false detection. However, our models managed to identify the vehicle and successfully differentiated between the person and vehicle. In the LISA dataset, labels were only provided for vehicles. Therefore, the proposed model only detected the vehicle.

\begin{figure}[H]
    \begin{subfigure}[t]{0.5\textwidth}
        \includegraphics[width=0.88\textwidth]{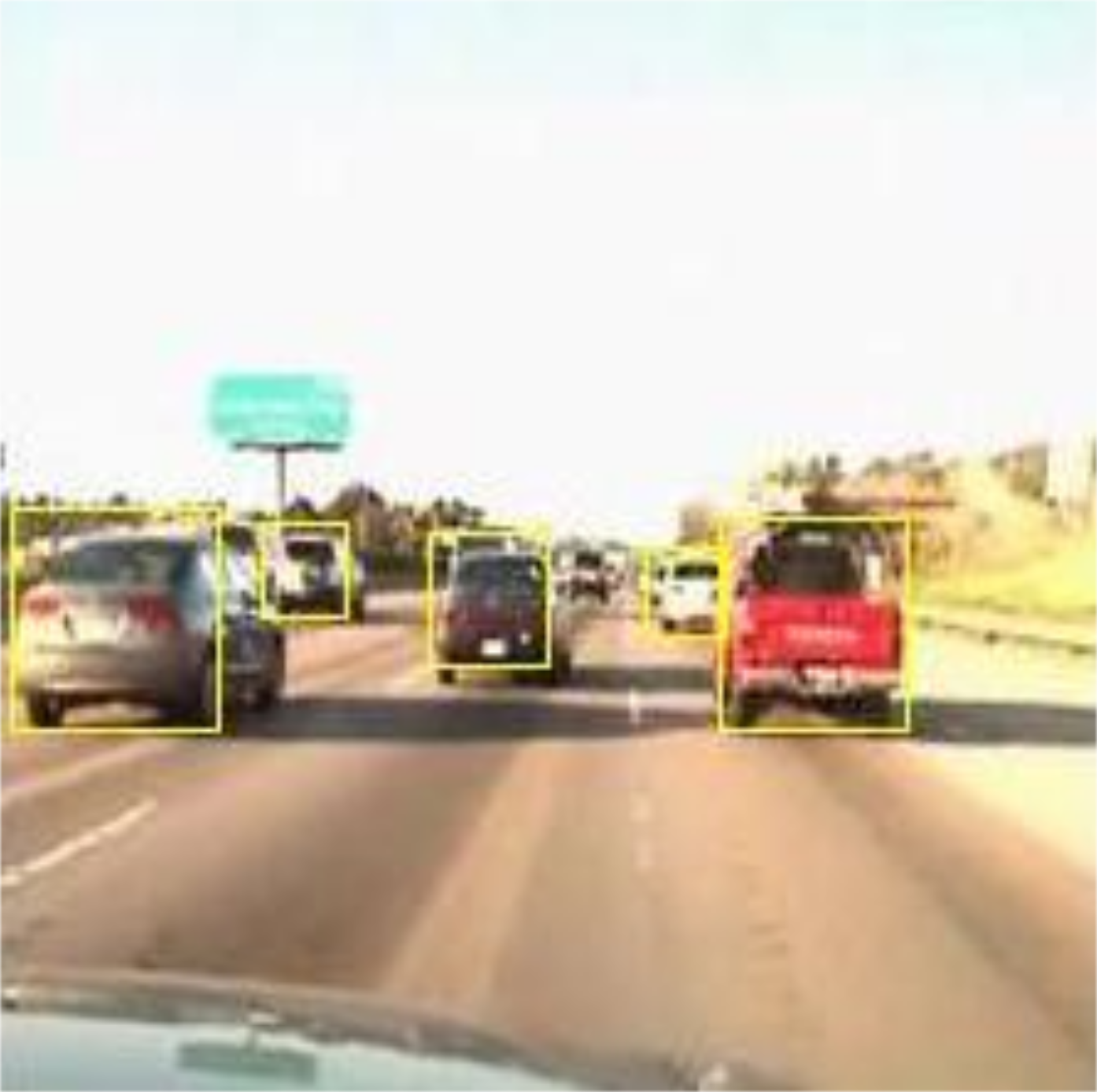}
        \caption{\centering} 
        \label{car1}
    \end{subfigure}
\vspace{6pt}
    \begin{subfigure}[t]{0.5\textwidth}
        \includegraphics[width=0.88\textwidth]{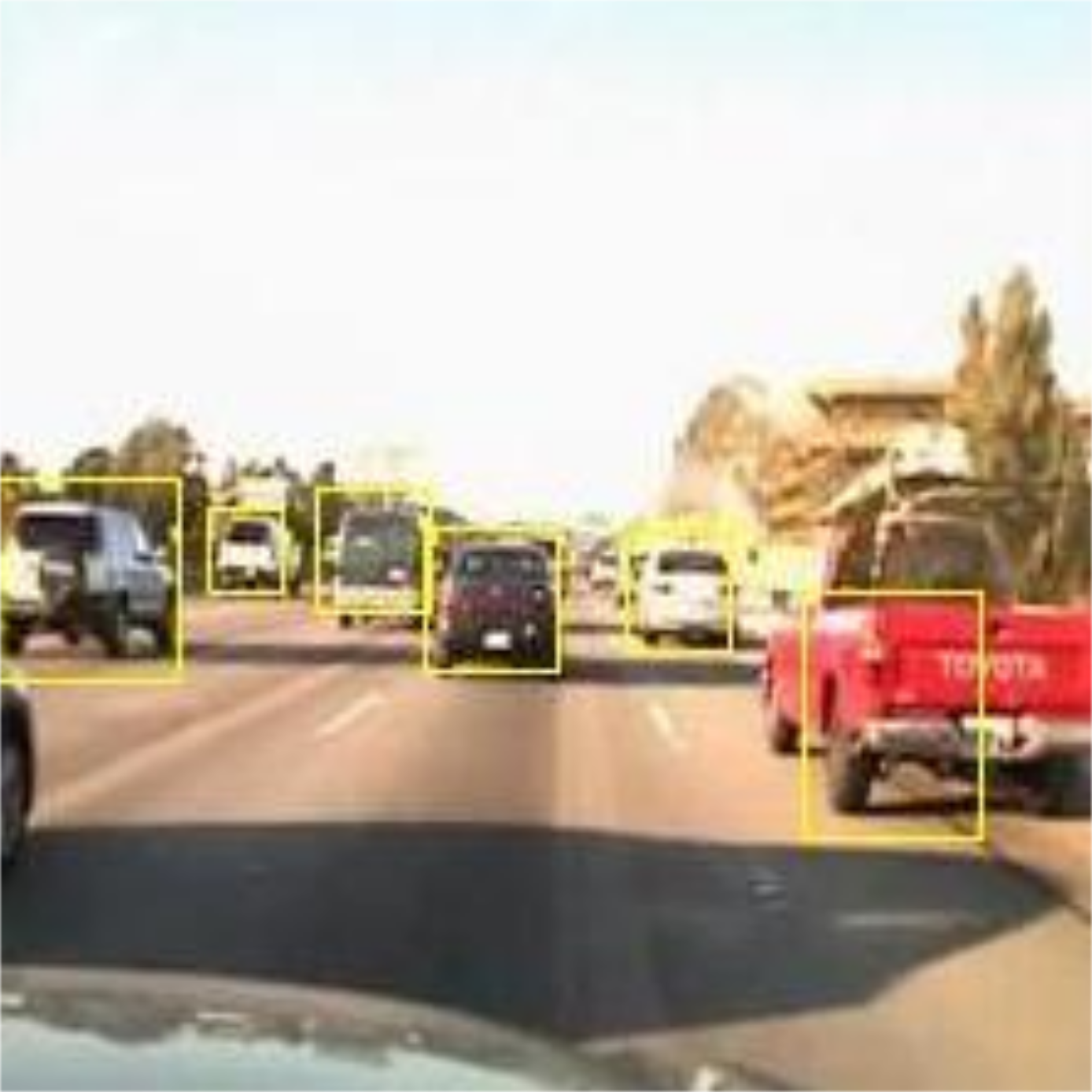}
        \caption{\centering} 
        \label{car2}
    \end{subfigure}
\vspace{6pt}
    \begin{subfigure}[t]{0.5\textwidth}
        \includegraphics[width=0.88\textwidth]{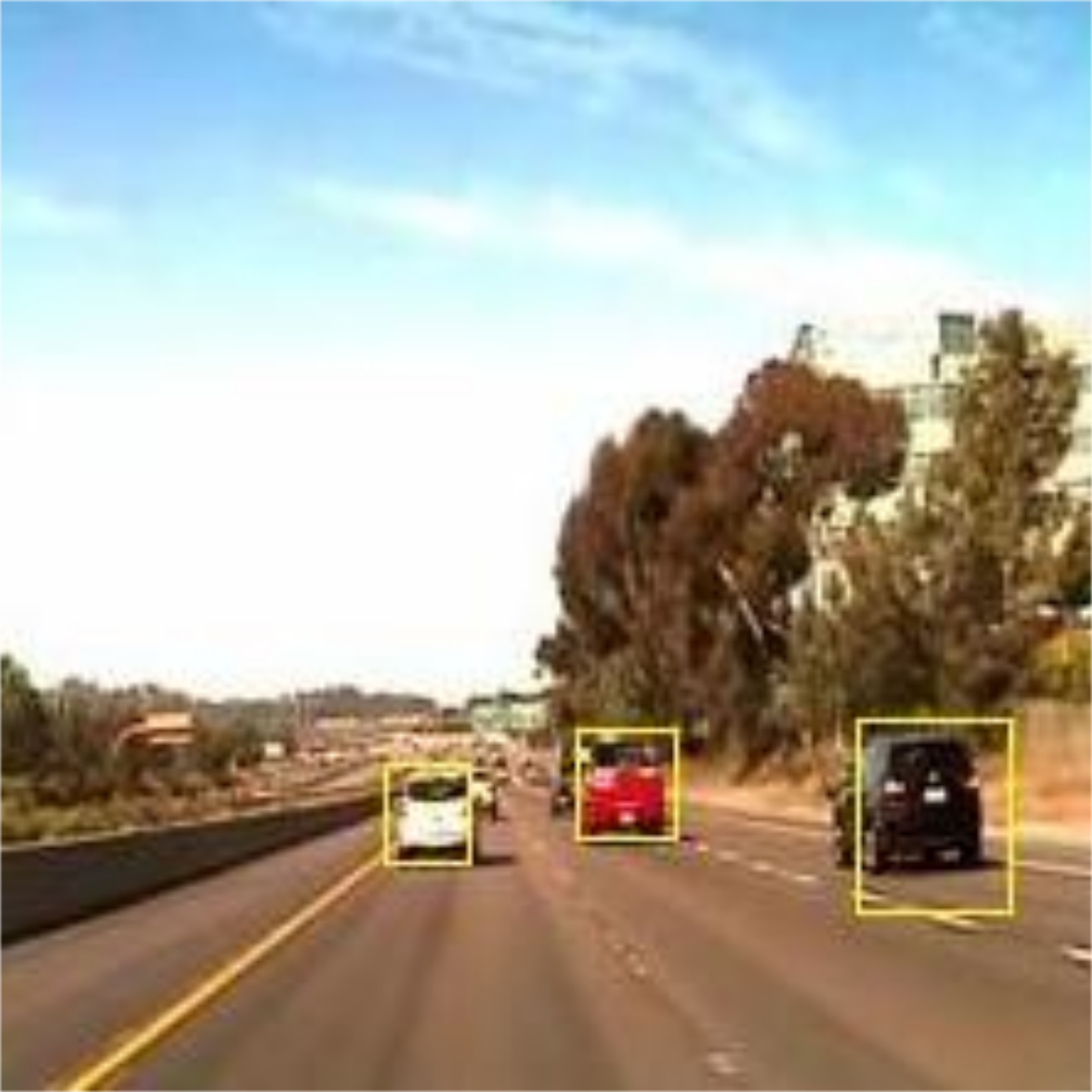}
        \caption{\centering} 
        \label{car3}
    \end{subfigure}
\vspace{6pt}
    \begin{subfigure}[t]{0.5\textwidth}
        \includegraphics[width=0.88\textwidth]{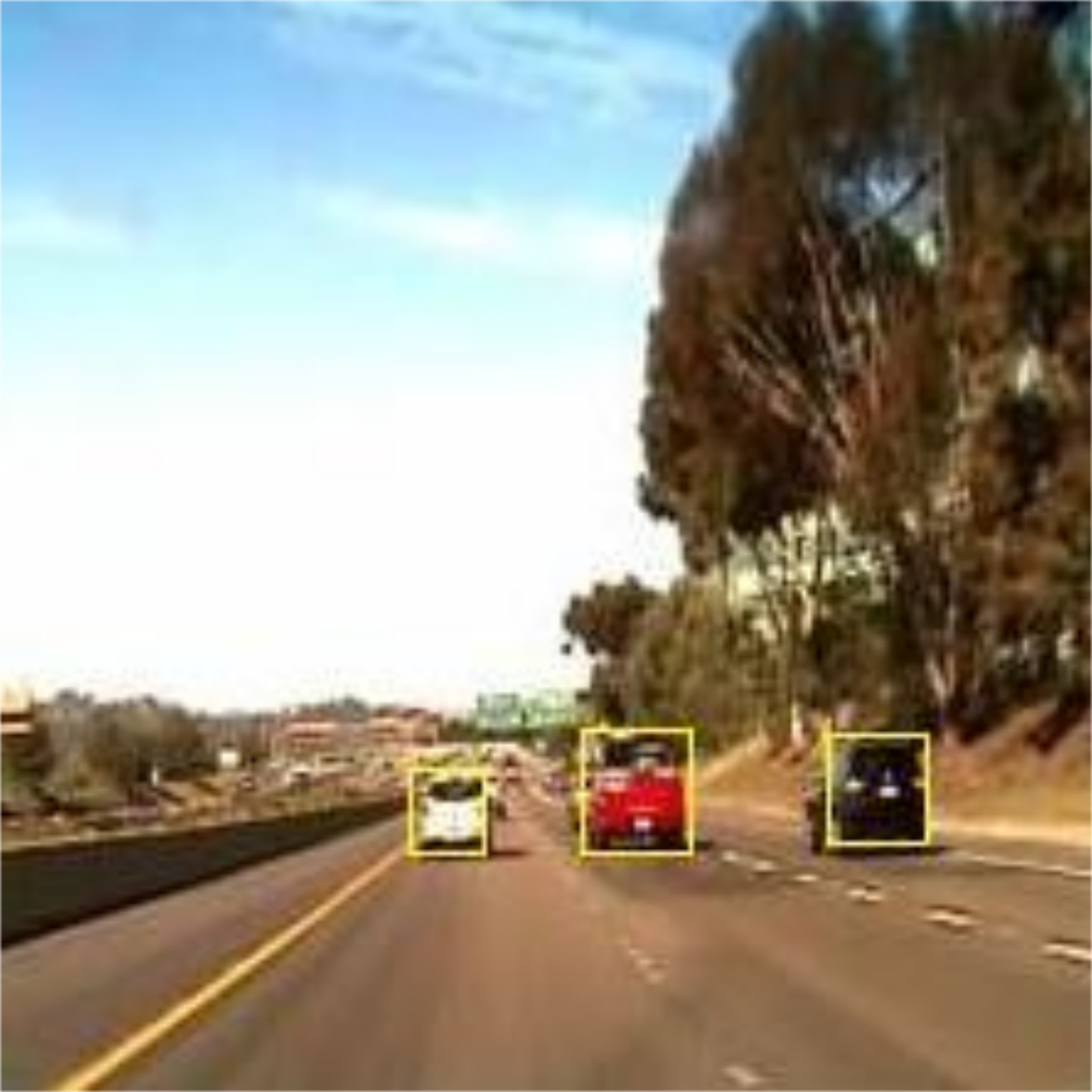}
        \caption{\centering} 
        \label{car4}
    \end{subfigure}
\vspace{6pt}
\caption{\textit{Cont.}}
\end{figure}
\begin{figure}[H]\ContinuedFloat
    \begin{subfigure}[t]{0.5\textwidth}
        \includegraphics[width=0.88\textwidth]{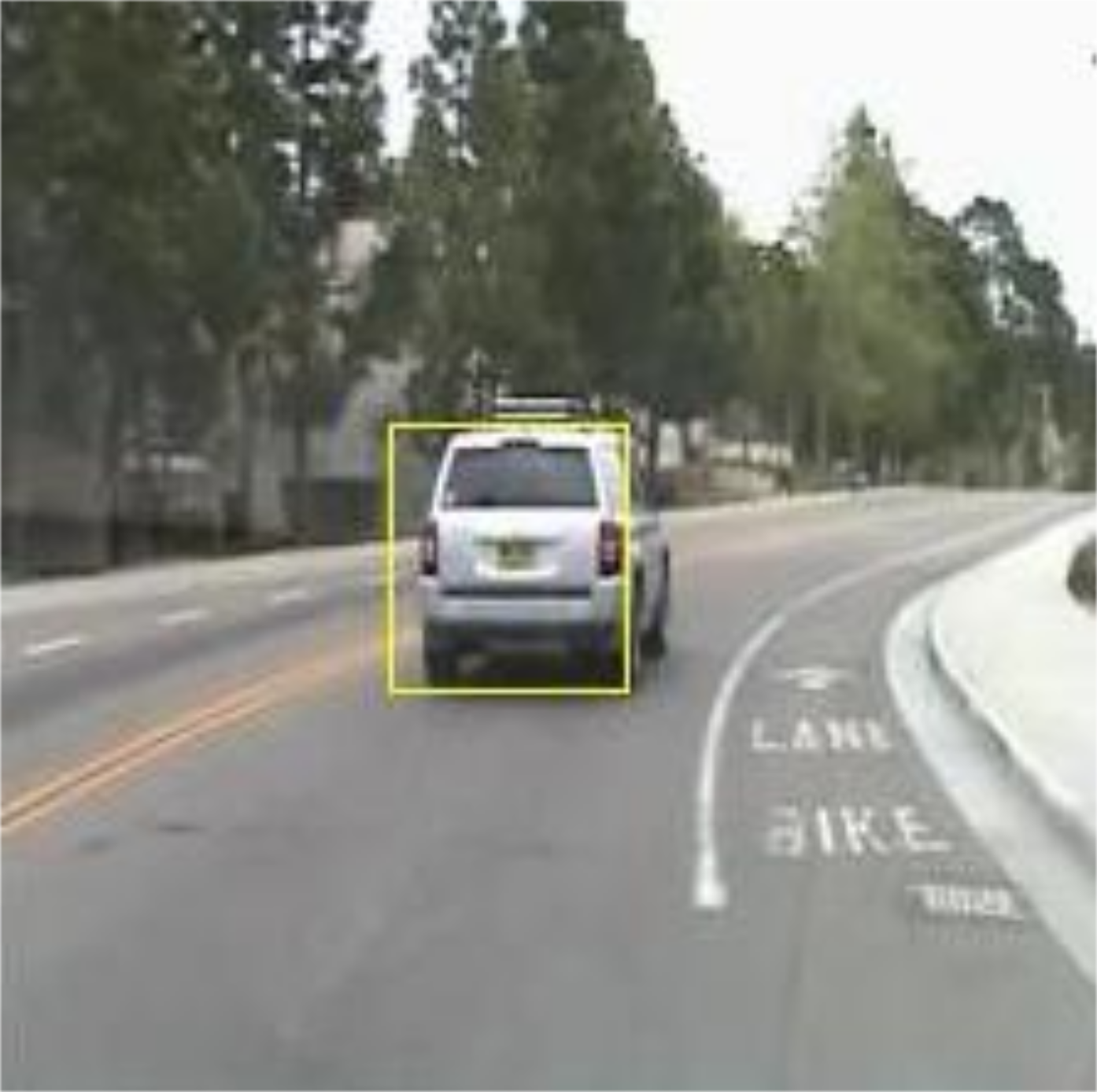}
        \caption{\centering} 
        \label{car5}
    \end{subfigure}
\vspace{6pt}
    \begin{subfigure}[t]{0.5\textwidth}
        \includegraphics[width=0.88\textwidth]{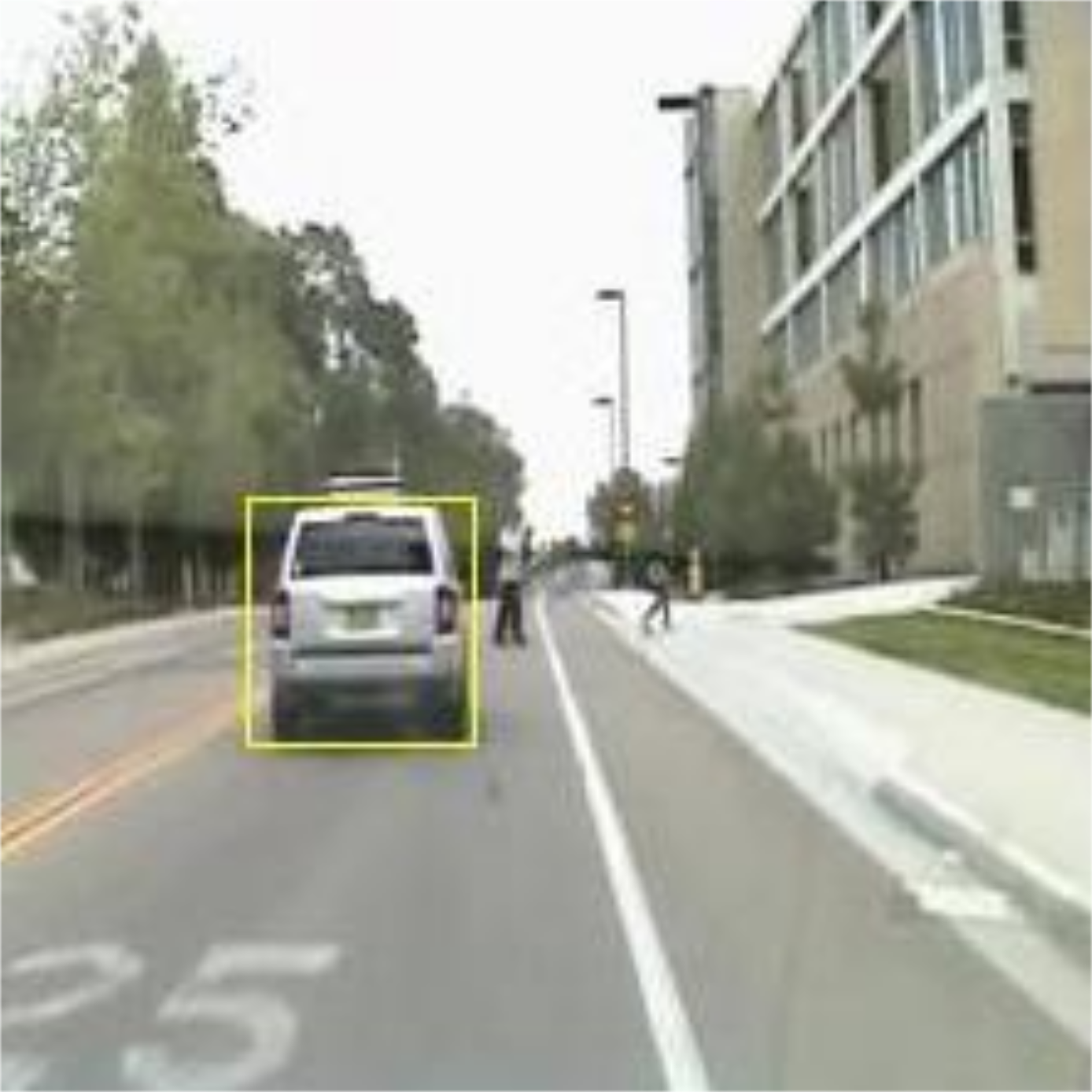}
        \caption{\centering} 
        \label{car6}
    \end{subfigure}
\caption{\textls[-5]{Vehicle detection from the online LISA dataset: (\textbf{a}) five vehicles detected in dense traffic scenario; (\textbf{b}) six vehicles detected in a dense traffic condition; (\textbf{c}) vehicle detection on highway; (\textbf{d})~vehicle detection on highway; (\textbf{e}) vehicle detection in urban area; and (\textbf{f}) differentiating vehicle and~pedestrian.} \label{car}}
\end{figure}

\begin{figure}[H]
\includegraphics[width=12cm]{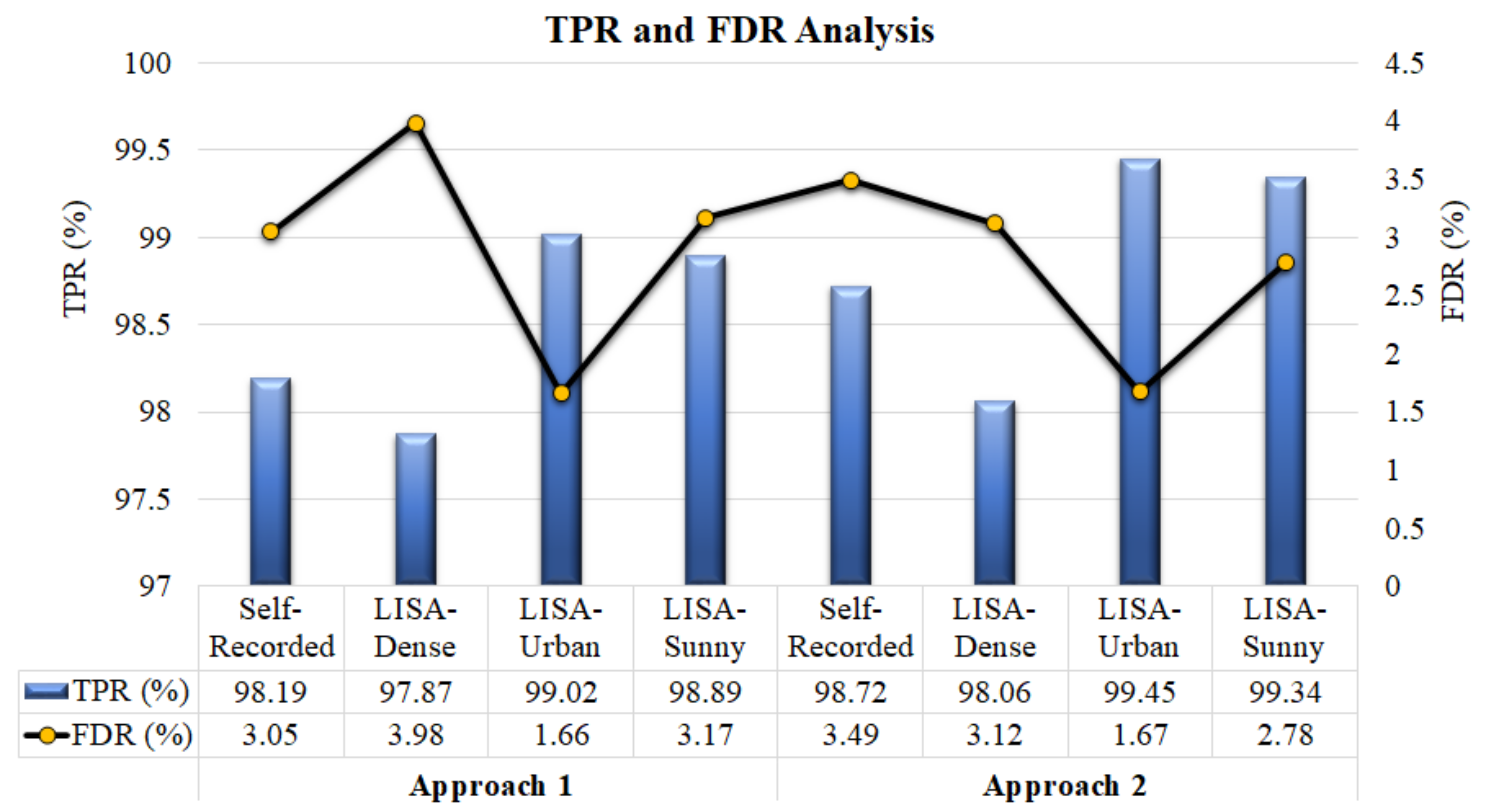}
\caption{TPR (\%) and FDR (\%) analysis of proposed approaches for self-recorded and online LISA~datasets. \label{results1}}
\end{figure}

The visual analysis of the true positive rate (TPR) and false detection rate (FDR) for proposed approaches against different sets of data is presented in Figure \ref{results1}; the figure shows that both approaches delivered reliable outcomes for the self-recorded as well as online datasets. The TPR value obtained from the faster R-CNN with pre-trained fused (Resnet 101 and Resnet 50) high-level feature descriptors is slightly higher compared to the the faster R-CNN with the proposed fused (DConNet and VeDConNet) feature descriptors. However, the faster R-CNN with the proposed feature descriptors provides a lower value of FDR for the self-recorded dataset and gives a comparable FDR for the LISA-Urban~dataset.

The frame rate (frames per second) for each type of dataset used in both approaches is given in Table \ref{tabframe}, which shows that the first model has a comparatively better frame rate. The pre-trained model (i.e., faster R-CNN with high-level feature descriptors of ResNet 101 and ResNet 50) took more time to compute features compared to the model presented in the first approach (i.e., faster R-CNN with high-level feature descriptors of DConNet and VeDConNet). Hence, the model presented in first approach is capable of providing significant performance for the vehicle detection scenarios where less computational time is required.

\begin{table}[H] 
\caption{\textls[-25]{Analysis of both models in terms of frame rate to validate the proposed deep neural architectures.}\label{tabframe}}
\newcolumntype{C}{>{\centering\arraybackslash}X}
\begin{tabularx}{\textwidth}{CCC}
\toprule
\textbf{Proposed Models}	& \textbf{Dataset}	& \textbf{Frame Rate (fps)} \\
\midrule
Model 1	& Self-Recorded	&   1.03\\
	            & LISA-Dense    &	1.10\\
	            & LISA-Urban    &	1.39\\
	            &  LISA-Sunny   &	1.14\\
\midrule	            
Model 2    & Self-Recorded &	0.89\\
	            & LISA-Dense    &	0.94\\
	            & LISA-Urban    &	1.12\\
	            & LISA-Sunny    &	1.00\\
\bottomrule
\end{tabularx}
\end{table}

\begin{figure}[H]
\includegraphics[width=13cm]{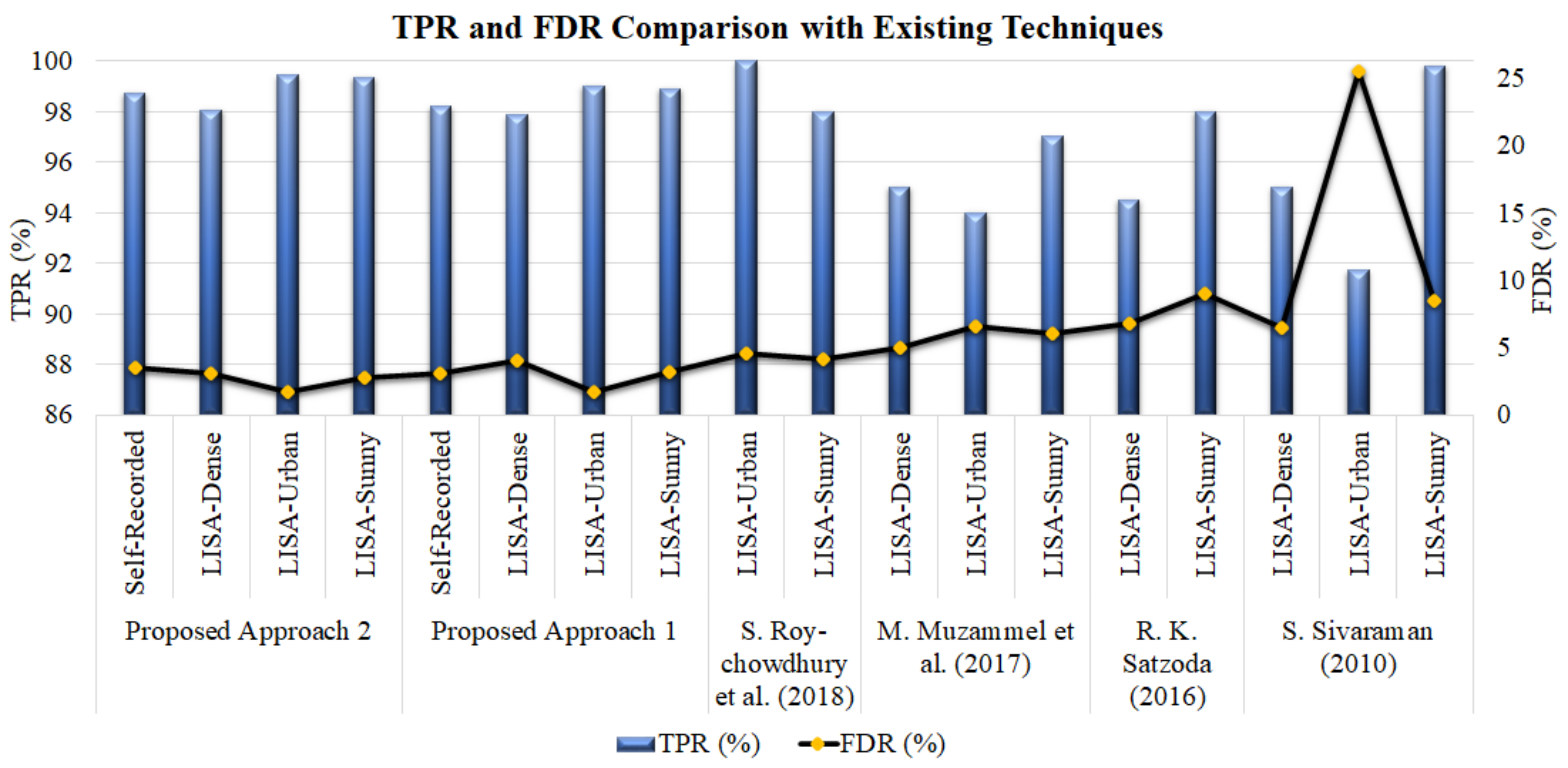}
\caption{TPR (\%) and FDR (\%) analysis of proposed experiments for self-recorded and online LISA~datasets.} \label{results2}
\end{figure}

The detailed comparisons of different parameters, including TPR, FDR, and frame rate from the existing state-of-the-art techniques and our proposed models, are presented in Table \ref{tabresults}. In addition, the graphical representation of true positive and false detection rates (i.e., TPR and FDR) of both models and their comparisons with the existing state-of-the-art approaches are given in Figure \ref{results2}.

\begin{table}[H] 
\caption{Comparisons of proposed approaches with existing state-of-the-art approaches in terms of true positive rate (TPR), false detection rate (FDR), and frame rate.\label{tabresults}}
\setlength{\cellWidtha}{\textwidth/5-2\tabcolsep--0.2in}
\setlength{\cellWidthb}{\textwidth/5-2\tabcolsep-0in}
\setlength{\cellWidthc}{\textwidth/5-2\tabcolsep-0.1in}
\setlength{\cellWidthd}{\textwidth/5-2\tabcolsep-0.3in}
\setlength{\cellWidthe}{\textwidth/5-2\tabcolsep--0.2in}
\scalebox{1}[1]{\begin{tabularx}{\textwidth}{>{\centering\arraybackslash}m{\cellWidtha}>{\centering\arraybackslash}m{\cellWidthb}>{\centering\arraybackslash}m{\cellWidthc}>{\centering\arraybackslash}m{\cellWidthd}>{\centering\arraybackslash}m{\cellWidthe}}
\toprule
\textbf{Reference} & \textbf{Dataset} & \textbf{TPR (\%)} 
& \textbf{FDR (\%)} & \textbf{Frame Rate (fps)}\\
\midrule
Proposed                   & Self-Recorded &	98.72 &	3.49 &	0.89\\
Approach 2               & LISA-Dense    &	98.06 &	3.12 &	0.94\\
	                       & LISA-Urban    &	99.45 &	1.67 &	1.12\\
	                       & LISA-Sunny    &	99.34 &	2.78 &	1.00\\
\midrule	
Proposed 	               & Self-Recorded &	98.19 &	3.05 &	1.03\\
Approach 1	           & LISA-Dense    &	97.87 &	3.98 &	1.10\\
	                       & LISA-Urban    &	99.02 &	1.66 &	1.39\\
	                       & LISA-Sunny    &	98.89 &	3.17 &	1.14\\
\midrule	
S. Roychowdhury             & LISA-Urban    &	100.00 & 4.50 &	1.10\\
et al. (2018) \cite{ref37} & LISA-Sunny    &	98.00 &	4.10 &	1.10\\
\midrule
M. Muzammel                 & LISA-Dense    &	95.01 & 5.01 & 29.04\\ 
et  al. (2017) \cite{ref36} & LISA-Urban    &	94.00 &	6.60 & 25.06\\ 
                            & LISA-Sunny    &	97.00 &	6.03 & 37.50\\ 
\midrule	
R. K. Satzoda              & LISA-Dense    &	94.50 &	6.80 & 15.50\\ 
(2016) \cite{ref38}       & LISA-Sunny	   &    98.00 &	9.00 & 25.40\\ 
\midrule	
S. Sivaraman               & LISA-Dense	   &    95.00 &	6.40 &	---\\ 
(2010) \cite{ref35}	   & LISA-Urban	   &    91.70 &	25.50 &	---\\ 
	                       & LISA-Sunny	   &    99.80 & 8.50 &	---\\ 
\bottomrule
\end{tabularx}}
\end{table}

From Table \ref{tabresults}, it can be deduced that our model achieved significantly higher results compared to the existing methods (deep learning and machine learning models). The deep learning model presented by S. Roychowdhury et al. (2018) \cite{ref37} was able to achieve 100\% and 98\% TPR for LISA-Urban and LISA-Sunny datasets, respectively. The proposed model (i.e., faster R-CNN with high-level feature descriptors of DConNet and VeDConNet) was able to get a higher TPR value for the LISA-Sunny dataset but a significantly close TPR value for the LISA-Urban dataset. Our model outperformed all the existing methods in terms of FDR. A very low false detection rate was obtained for all three online datasets (LISA-Dense, LISA-Sunny, and LISA-Urban) compared to the existing machine/deep learning techniques. Moreover, higher TPR values have been acquired for all three LISA datasets compared to the existing machine learning techniques.

From Figure \ref{results2}, one can see that, for the first model, the FDR is less than 4\% for all datasets, making it suitable for real time applications. Further, TPR values are almost constant for all types of datasets. It shows that the model achieved a reliable result for all types of scenarios. 

\subsection{Discussion}

The proposed approaches were successfully able to detect different types of vehicles, such as motorcycles, cars, trucks, etc. In addition, both approaches proved to be reliable for dense traffic conditions for the online LISA dataset. The fusion of pre-trained networks provided a higher accuracy for both the self-recorded and online datasets compared to the first approach in which two self designed CNNs are used. However, for the first approach, a lower frame rate was obtained compared to the second approach.      

For the online datasets, both approaches obtained either a high or comparable accuracy compared to the existing state-of-the-art approaches, as given in Table \ref{tabresults}. For LISA-Dense, the highest TPR value of 98.06\% was obtained by the second proposed approach, followed by the first approach with a value of 97.87\%. Further, the machine learning approaches proposed by M. Muzammel et al. (2017) \cite{ref36}, R. K. Satzoda (2016) \cite{ref38}, and S. Sivaraman~\cite{ref35} (2010) reported TPR values of 95.01\%, 94.50\%, and 95\%, respectively. \mbox{S. Roychowdhury et~al.} (2018) \cite{ref37} did not report any results for the LISA-Dense dataset. For LISA-Urban, the highest TPR value was obtained by S. Roychowdhury et al. (2018)~\cite{ref37}, followed by the proposed second approach. For LISA-Urban, the lowest TPR value of 91.70\% was obtained by S. Sivaraman \cite{ref35}.  

From Figure \ref{results2}, the fusion of features significantly improved the performance of faster R-CNN. A notable reduction in false detection was found for the online datasets compared to the deep learning \cite{ref37} and machine learning approaches \cite{ref35,ref36,ref38}. A system with lower false detection rate will provide fewer false warnings and thus increase the trust of the drivers for the system. It has been found in the literature that collision warnings reduce the attention resources required for processing the target correctly \cite{muzammel2018event}. In addition, collision warnings facilitate the sensory processing of the target \cite{fort2013impact,bueno2012electrophysiological}. Finally, our fusion technique results are in line with studies in \cite{yang2017multimodal, MUZAMMEL2021106433,mendels2017hybrid}.

With regard to the comparison between both approaches, the model presented in the first approach obtained a lower FDR compared to the model presented in the second approach for the self-recorded and LISA-Urban datasets. In addition, the model presented in the first approach has a higher frame rate for all the datasets compared to the model presented in the second approach. In other TPR and FDR values, the second approach model outperformed the first approach model. Therefore, there is a slight trade off between performance and computation time.

\section{Conclusions and Future Work}
In this research, we propose deep neural architectures for blind-spot vehicle detection for heavy vehicles. Two different models for feature extraction are used with the faster R-CNN network. Furthermore, the high-level features obtained from both networks are fused together in order to improve the network performance. The proposed models successfully detected blind-spot vehicles with reliable accuracy using both the self-recorded and publicly available datasets. Moreover, the fusion of feature extraction networks improved the results significantly, and a notable increment in performance is observed. In addition, we compared our fusion model with the state-of-the-art benchmark, machine learning, and deep learning approaches. Our proposed work outperformed all the existing approaches for vehicle detection in various scenarios, including dense traffics, urban surroundings, with and without pedestrians, shadows, and different weather conditions. The proposed model is capable enough to be usedfor  not only buses but also other heavy vehicles such as trucks, trailers, oil tankers, etc. This research work is limited to the integration of only two convolutional neural networks with faster R-CNN. In the future, more than two convolutional neural networks may be integrated with faster R-CNN, and parametric study for accuracy and frame rate may be performed. 

\vspace{6pt} 



\authorcontributions{Conceptualization, M.Z.Y.; data curation, M.M. and M.N.M.S.; formal analysis, M.A.A.; investigation, M.N.M.S.; methodology, M.M.; supervision, M.Z.Y.; validation, F.S.; visualization, M.A.A.; writing---original draft, M.M. and F.S.; writing---review and editing, M.Z.Y., M.N.M.S., F.S. and M.A.A. All authors have read and agreed to the published version of the manuscript.}

\funding{{This} 
 research was supported in part by Ministry of Education Malaysia under Higher Institutional Centre of Excellence (HICoE) Scheme awarded to the Centre for Intelligent Signal and Imaging Research (CISIR), Universiti Teknologi PETRONAS (UTP), Malaysia; and, in part, by the Yayasan Universiti Teknologi PETRONAS (YUTP) Fund under Grant 015LC0-239.}

\institutionalreview{Not applicable.}

\informedconsent{Not applicable.}


\dataavailability{Not applicable.}

\acknowledgments{We express our gratitude and acknowledgment to the Centre for Intelligent Signal and Imaging Research (CISIR) and Electrical and Electronic Engineering Department, Universiti Teknologi PETRONAS (UTP), Malaysia.}

\conflictsofinterest{The authors declare no conflict of interest.} 



\abbreviations{Abbreviations}{
The following abbreviations are used in this manuscript:\\

\noindent 
\begin{tabular}{@{}ll}
CNN & Convolutional neural network\\
HOG &  Histogram of oriented gradients\\
SS & Selective search \\
YOLO & You only look once \\
LISA & Laboratory for Intelligent and Safe Automobiles \\ 
SGDM & Stochastic gradient descent with momentum\\
TPR & True positive rate\\ 
FDR  & False detection rate
\end{tabular}
}

\begin{adjustwidth}{-\extralength}{0cm}

\reftitle{References}




%
%
%
\end{adjustwidth}
\end{document}